\documentclass[11pt,a4paper]{article}
\usepackage{acl2015}
\usepackage{times}
\usepackage{url}
\usepackage{latexsym}
\usepackage{tabularx} % in the preamble
\usepackage[table,xcdraw]{xcolor}
\usepackage{enumitem}
\usepackage{graphicx}

\title{Exposing Paid Opinion Manipulation Trolls}

\author{Todor Mihaylov, Ivan Koychev\\
  FMI\\
  Sofia University\\
  {\tt \small tbmihailov@gmail.com} \\
    {\tt \small koychev@fmi.uni-sofia.bg} \\
\\\And
 Georgi D. Georgiev \\
  Ontotext AD \\
  Sofia, Bulgaria\\
  {\tt \small georgiev@ontotext.com}  \\\And
   Preslav Nakov \\
  Qatar Computing Research Institute\\
  HBKU, Qatar \\
  {\tt \small pnakov@qf.org.qa} 
  \\}

\date{}

\begin{document}
\maketitle
\begin{abstract}
Recently, Web forums have been invaded by \emph{opinion manipulation trolls}. Some trolls try to influence the other users driven by their own convictions, while in other cases 
%they could be paid by a government or a PR agency.
they can be organized and paid, e.g.,~by a political party or a PR agency that gives them specific instructions what to write.
%This paper describes our work on opinion manipulation troll detection in news community forums. 
%We do our research on data from Bulgarian forums. 
Finding paid trolls automatically using machine learning is a hard task, as there is no enough training data to train a classifier;
%. This blocks research as it makes it hard to obtain good training data; 
yet some test data is possible to obtain, as these trolls are sometimes caught and widely exposed.
%We collect troll and non-troll profiles and we generate some statistical features that model troll behavior based on several assumptions and observations about trolls, coming from various media publications about internet trolling. 
%We have been lucky to have the IDs of 15 known paid trolls, a subset of which we could use for testing. For training, 
In this paper, we solve the training data problem by assuming that a user who is called a \emph{troll} by several different people is likely to be such, and one who has never been called a troll is unlikely to be such.
%We call these trolls ``mentioned'' trolls. 
%Our main goal is to identify real paid trolls so we train a classifier with such ``mentioned'' trolls and evaluate with known paid trolls and we achieve some good results.
%We experiment with different variations of this definition, and in each case we show that we can train a classifier to distinguish a likely troll from a non-troll with very high accuracy, 85--98\%, thanks to our rich feature set.
We compare the profiles of (\emph{i})~paid trolls vs. (\emph{ii})~``mentioned'' trolls vs. (\emph{iii})~non-trolls, and we further show that a classifier trained to distinguish (\emph{ii}) from (\emph{iii}) does quite well also at telling apart (\emph{i}) from (\emph{iii}).
%We experiment with different variations of this ``mentioned'' troll definition, and in each case we show how close is it to paid trolls depending on the evaluation results.
%In our work, we train a classifier with equal number of positive and negative primers and we test with several feature groups configurations. Our classifier performs convenient and prove and reject some of our assumptions. 
\end{abstract}

\section{Introduction}
\label{sec:intro}

% With the rise of social media it became normal for people to read and follow other users' opinion. This created the opportunity for corporations, governments and others to distribute rumors, misinformation, speculation and to use other dishonest practices to manipulate user opinion \cite{derczynski2014pheme}. They could consistently use trolls \cite{cambria2010not}, write fake posts and comments in public forums, thus making veracity one of the challenges in digital social networking \cite{derczynski2014spatio}. 

During the 2013-2014 Bulgarian protests against the Oresharski cabinet, social networks and news community forums became the main ``battle grounds'' between supporters and opponents of the government.
In that period, there was notable censorship in the media, and many people who lived outside the capital did not really know what was actually happening. Moreover, there was a very notable presence of government supporters in Web forums. 
In series of leaked documents in the independent Bulgarian media Bivol,\footnote{\scriptsize \url{https://bivol.bg/en/category/b-files-en/b-files-trolls-en}} it was alleged that the ruling Socialist party was paying Internet trolls with EU Parliament money.

The Bivol's leaked documents revealed for the first time such a practice by a political party despite the problem with opinion manipulation being 
generally 
notable across Eastern Europe.
The reputation management documents described the following services:\textit{``Monthly posting online of 250 comments by virtual users with varied, typical and evolving profiles from different (non-recurring) IP addresses to inform, promote, balance or counteract. The intensity of the provided online presence will be adequately distributed and will correspond to the political situation in the country.''}

The practice of using Internet trolls for opinion manipulation has been reality since the rise of Internet and community forums. It has been shown that user opinions about products, companies and politics can be influenced by opinions posted by other online users %in online forums and social networks 
\cite{Dellarocas06}. This makes it easy for companies and political parties to gain popularity by paying for ``reputation management'' to people that write in discussion forums and social networks fake opinions from fake profiles. 
%In Europe, the problem has emerged in the last few years, especially in the context of the Ukraine-Russia conflict;
%which started at the end of 2013,
%this was also discussed in international media
%\footnote{\url{http://www.forbes.com/sites/peterhimler/2014/05/06/russias-media-trolls/}},  \footnote{\url{http://www.theguardian.com/commentisfree/2014/may/04/pro-russia-trolls-ukraine-guardian-online}}.%\cite{Forbes-Rusia-Trolls,Guardian-Russia-Trolling}. 
Yet, over time, forum users developed sensitivity about trolls, and started publicly exposing them. 
%Still, it is hard for forum administrators to block them as trolls try formally not to violate the forum rules. 
%Over time, forum users developed a sensitivity for political trolling and some can easily recognize troll comments and profiles.
%Reading more and more about trolling in news and discussions most users developed immunity in political trolling and can easily recognize troll and non-troll comments and profiles. Even more - many users are accusing other users of being trolls. However, it is not easy for the forum administrators to identify and block internet trolls as they are real people and they try not to formally violate the forum rules. 

\section{Related Work}
\label{sec:related}

A popular way to manipulate public opinion in Intternet is by making controversial posts on a specific topic that aim to win the argument at any cost, usually accompanied by untruthful and deceptive information. The problem of deceptive opinion spam is studied in \cite{Ott:2011:FDO:2002472.2002512}, where the authors integrated work from both psychology and computational linguistics trying to detect fake opinions that were written to sound authentic. Malicious troll users posting misinformation posts have also been studied using graph-based approaches over signed social networks \cite{Ortega20122884,kumar2014accurately}. A related problem is that of trustworthiness of statements on the Web \cite{rowe2009assessing}.

Troll detection and offensive language use are understudied problems \cite{xu2010filtering}. They have been addressed using analysis of the semantics and the sentiment in posts
~%to filter out trolls 
\cite{cambria2010not}; there have been also studies of general troll behavior \cite{herring2002searching,buckels2014trolls}. Another approach has been to use lexico-syntactic features about user's writing style, structure, and cyber-bullying content 
%as features to predict probability to send out offensive content
\cite{chen2012detecting}; cyber-bullying was detected using user profile and post metadata \cite{Garcia2013}, and sentiment analysis \cite{Xu:2012:FLS}.   

A related problem is that of Web spam detection, 
usually addressed as text classification \cite{sebastiani2002machine}, e.g., using spam keyword spotting \cite{dave2003mining}, lexical affinity of arbitrary words to spam content \cite{hu2004mining}, frequency of punctuation and word co-occurrence \cite{li2006combining}.
See \cite{Castillo:2011:AWS} for an overview on adversarial Web search.

%\cite{Dellarocas06}

\section{Data}
\label{sec:data}

We crawled the largest media community forum in Bulgaria, that of Dnevnik.bg\footnote{\url{http://dnevnik.bg}}, a daily newspaper that requires users to be signed in to  comment (all in Bulgarian), which makes it easy to track them.
%Typical for this media is that it posts news publications in various categories on a daily basis and it requires users be signed in to read and comments news.
%This makes it easy to identify users that post comments regularly.
%In the forum official language is Bulgarian and all comments are written in Bulgarian.
The platform allows users to comment on news, to reply to other users' comments and to vote on them with thumbs up/down. Each publication has a category, a subcategory, and a list of manually selected tags (keywords).
%it could be assigned to a specific topic. 

We crawled 
%all publications in 
the \emph{Bulgaria}, \emph{Europe}, and \emph{World} categories
%which are mostly about politics,
for the period 01-Jan-2013 to 01-Apr-2015, together with the comments and the corresponding user profiles:
%that have written at least 1 comment. In our dataset, we have 
34,514 publications on 232 topics and with 13,575 tags, 1,930,818 comments (897,806 of them replies), and 14,598 users.

We have three groups of users: known paid trolls (as exposed in Bivol), ``mentioned'' trolls (called trolls by a certain number of different users), and non-trolls (never called trolls, despite having a high number of posts).
%The first group contains 15 known paid troll users exposed in a leaked reports of troll work results.% \cite{bivol.bg-yotova,Bivol.bg-trollogie}.
%We also have  \emph{trolls} users who were called such by five or more distinct users, and \emph{non-trolls} if they have never been called so. 
Looking at users with at least 150 comments, we have 314 ``mentioned'' trolls (mentioned by five or more users) vs. 964 non-trolls (vs. some in between); we further have 15 paid trolls from Bivol. Here is an example post with troll accusation (translated):
%of real comment replies translated (in English):

%\textit{``To comment from "Historama": Murzi \texttt{(troll in Russian)}, you know that you cannot manipulate public opinion, right?''} 

\textit{``To comment from "Rozalina": You, \underline{trolls}, are so funny :) I saw the same signature under other comments:)''}

% \begin{table}[tbh]
% \centering
% \vspace{-1pt}
% \caption{Dataset statistics.}
% \label{table:dataset-statistics}
% \vspace{-1pt}
% \label{Data statistics}
% \begin{tabular}{lr}
% %\hline 
% \textbf{Object}          & \textbf{Count}   \\  \hline 
% Publications       & 34,514   \\ 
% Comments           & 1,930,818 \\
%   -of which replies& 897,806 \\
% User profiles      & 14,598   \\
% Topics             & 232     \\ 
% Tags               & 13,575   \\  \hline
% \end{tabular}
% \vspace{-1pt}
% \end{table}

% \begin{table}[tbh]
% \centering
% \caption{Number of known paid trolls, ``mentioned'' troll (by at least 5 users) and non-troll profiles with min 100 comments}
% \label{table:user-profiles-statistics}
% \begin{tabular}{lc}
% %\hline 
% \textbf{Type}          & \textbf{Count}   \\  \hline 
% Known paid troll users : (all)     & 15   \\ 
% ``Mentioned'' troll users: (all)     & 314   \\ 
% Non-troll users: (all)   & 964 \\\hline    
% Non-troll users: (selected)   & 314 \\\hline    
% \end{tabular}
% \end{table}

\section{Method}
\label{sec:method}

%In our current work we are using only non-language specific features so we can model troll users non-language specific  behavior. 
%
%\subsection{Motivation}
We train a classifier to distinguish ``mentioned'' trolls vs. non-trolls; we experiment both with balanced and (natural) imbalanced classes. Then, at test time, we evaluate how well the classifier performs at discriminating paid trolls vs. non-trolls.
We use a support vector machine (SVM) classifier \cite{libsvm} with a radial basis function (RBF) kernel,
and features motivated by several publications about troll behavior. 
%For each user profile, we extract base statistics features that are motivated by a specific troll user behavior. 
%
%
%\subsection{Features}
%

Note that we perform the classification at the user level, i.e., based on user activity history, from which we extract statistics summarizing the user activity.
In particular, for each user, we count the number of comments posted, the number of days in the forum, the number of days with at least one comment, and the number of publications commented on. All other features are scaled with respect to these statistics, which makes it 
%for every user which make it 
possible for us to handle users that registered only recently (which we need to do at test time). Our features can be divided in the following general groups:

\textbf{Vote-based features.} We calculate the number of comments with positive and negative votes for each user. This is useful as we assume that non-trolls are likely to disagree with trolls, and to give them negative votes. We use the sum from all comments as a feature. We also count separately the comments with high, low and medium positive to negative ratio. Here are some example features:
%\textit{cmnt-vote-neg-high-cnt} is 
(a)~the number of comments where (positive$/$negative) $<$ 0.25, 
%\textit{cmnt-vote-neg-med-cnt} is 
and
(b)~the number of comments where (positive$/$negative) $<$ 0.50.
%, and so on. 

\textbf{Comment-to-publication similarity.}
These features measure the similarity between comments and publications.
We use cosine and TF.IDF-weighted vectors for the comment and for the publication. The idea is that trolls might try to change or blurr the topic of the publication if it differs from his/her views or agenda.

\textbf{Comment order-based features.} We count how many user comments the user has among the first $k$.
% comments under a publication. 
The idea is that trolls might try to be among the first to comment to achieve higher impact.
% (these comments stay on top of the other).

\textbf{Top loved/hated comments.} We calculate the number of times the user's comments were among the top 1, 3, 5, 10 most loved/hated comments in some thread. The idea is that in the comment thread below many publications there are some trolls that oppose all other users, and usually their comments are among the most hated.

\textbf{Comment replies-based features.} These are features that count how many comments by a given user are replies to other users' comments, how many are replies to other replies, and so on. The assumption here is that trolls post not only a large number of comments, but also a large number or replies, as they want to dominate the conversation, especially when defending a specific cause. We further generate complex features that combine user comment reply features and vote counts-based features, thus generating even more features that model the relationship between replies and user agreement/disagreement.

\textbf{Time-based features.} We generate features from the number of comments posted during different time periods on a daily or on a weekly basis. We assume that users who are paid or who could be activists of political parties probably have some usual times to post, e.g., maybe they do it as a full-time job.
% all day long, from the morning till the evening. 
On the other hand, most non-trolls work from 9:00 to 18:00, and thus we could expect that they should probably post less comments during this part of the day.
We have time-based features 
%like \textit{cmnt-hour-09, cmnt-hour-12}, which means 
that count the number of comments from 9:00 to 9:59, from 12:00 to 12:59, during working hours 9:00-18:00, etc.
%or \textit{cmnt-hour-work time}, which means 
%comments during work time 9:00-18:00, \textit{cmnt-dow-workdays}, and so on.

Note that all the above features are scaled, i.e.,~divided by the number of comments, by the number of days the user has spent in the forum, by the number of days in which the user posted more than one comment, etc.
%Other features are divided by each other in a reasonable manner. 
Overall, we have a total of 338 such scaled features.
In addition, we define a new set of features, which are non-scaled.

\textbf{Non-scaled features.} The non-scaled features are features based on the same statistics as above, but they are 
%This means that older users will be likely to have higher values for these non-scaled features. 
not divided by the number of comments / number of days in the forum / number of days with more that one comment, etc. 
For example, one non-scaled feature is
the number of times a comment by the target user was voted negatively, i.e., as thumbs down, by other users.
As a non-scaled feature, we would use this number directly, while above we would scale it by dividing it by the total number of user's comments,
by the total number of publications the user has commented on, etc.
Obviously, there is a danger in using non-scaled features: older users are likely to have higher values for them compared to recently-registered users. Yet, we found unscaled features useful in previous experiments \cite{mihaylov-georgiev-nakov:2015:CoNLL}, so we included them here as well.

\section{Experiments and Evaluation}
\label{sec:experiments}

In previous work \cite{mihaylov-georgiev-nakov:2015:CoNLL}, we have already experiments with distinguishing ``mentioned'' trolls vs. non-trolls, achieving accuracy of 88-94\%.
Here, we are interested in discriminating between \emph{paid} trolls and non-trolls. 

Unfortunately, we only know fifteen paid trolls (from the publication in Bivol), which is too little to use for training and testing. Thus, we trained on ``mentioned'' trolls vs. non-trolls, but we then tested on \emph{paid} trolls vs. non-trolls.
We focused on the top four known paid trolls with the highest number of posts, as they had more than 100 comments, which means that we had enough information about them.\footnote{There were six known paid trolls with more than 40 comments, and the remaining nine known paid trolls from Bivol had less than 40 comments.} Thus, for testing we used the four trolls with 100 posts or more, to which we added four non-trolls (i.e., users who have never been called \emph{trolls}).
For training, we used 314 ``mentioned'' troll with 150 posts or more, to which we added 314 non-trolls, also with 150+ posts.

For the experiments, we extracted the features described in the previous section, both scaled and non-scaled, and we normalized them in the -1 to 1 interval. We then trained a support vector machine (SVM) classifier \cite{libsvm} with a radial basis function (RBF) kernel with C=32 and g=0.0078125. We chose these values using cross-validation on the training dataset.
The testing results are shown in Tables \ref{table:train-mentioned-eval-bivol-all-features} and \ref{table:tr-ment-eval-bivol-some-features}. 

%For training we use balanced ``mentioned'' trolls and non-trolls.% with at least 150 comments. 

\begin{table*}[ht]
\centering
\begin{tabular}{lcccc}
\textbf{Features}                         & \textbf{Accuracy}& \textbf{Precision} & \textbf{Recall} & \textbf{F-score} \\\hline

All Scaled (AS)                  & 0.88     & 1.00      & 0.75   & 0.86    \\
AS - comment order (Scaled - S)  & 0.88     & 1.00      & 0.75   & 0.86    \\
AS - is reply (S)                & 0.88     & 1.00      & 0.75   & 0.86    \\
AS - is reply to has reply (S)   & 0.88     & 1.00      & 0.75   & 0.86    \\
AS - similarity (S)              & 0.88     & 1.00      & 0.75   & 0.86    \\
AS - similarity top (S)          & 0.88     & 1.00      & 0.75   & 0.86    \\
AS - topl oved hated (S)         & 0.88     & 1.00      & 0.75   & 0.86    \\
AS - total comments (S)          & 0.88     & 1.00      & 0.75   & 0.86    \\
AS - triggered replies range (S) & 0.88     & 1.00      & 0.75   & 0.86    \\
AS - triggered replies total (S) & 0.88     & 1.00      & 0.75   & 0.86    \\
AS - vote updown total (S)       & 0.88     & 1.00      & 0.75   & 0.86    \\\hline
AS - time (S)                    & 0.75     & 1.00      & 0.50   & 0.67    \\
AS - time hours (S)              & 0.75     & 1.00      & 0.50   & 0.67    \\
AS - vote up/down reply status (S) & 0.75     & 1.00      & 0.50   & 0.67    \\\hline
AS - time day of week (S)        & 0.63     & 1.00      & 0.25   & 0.40    \\
AS + Non Scaled (NS)             & 0.63     & 1.00      & 0.25   & 0.40    \\\hline
AS - vote up/down all (S)         & 0.38     & 0.00      & 0.00   & 0.00   \\\hline
\end{tabular}
\caption{
Results for classifying 4 paid trolls vs. 4 non-trolls for All Scaled (AS) `$-$' (minus) some scaled feature group. We train on 314 ``mentioned'' trolls vs. 314 non-trolls.
(The bottom features are better, as they yield the highest drop in accuracy and F1 when excluded from All Scaled.)}
\label{table:train-mentioned-eval-bivol-all-features}
\end{table*}

Table \ref{table:train-mentioned-eval-bivol-all-features} shows that we can find paid trolls with 100\% precision and 75\% recall, which is quite good.
However, we should be very cautious about any conclusions we draw, as we only had eight testing examples.
Yet, let us try to do some analysis.
First, note that the best F-score is achieved when using All Scaled features. 
Moreover, features based on reply status, similarity, up/down votes, number of triggered replies seem to have no impact on the classification performance, as excluding them from the All Scaled features does not affect the results either way. However, excluding time-related features and reply comments vote-based features results in bad score, which means that these features have the most impact on finding paid trolls. Finally, excluding all vote-related features results in zero precision and recall on paid trolls evaluation, which means that these features are key for finding paid trolls. 

\begin{table*}[!]
\centering
\begin{tabular}{lcccc}
\textbf{Features}                         & \textbf{Accuracy} & \textbf{Precision} & \textbf{Recall} & \textbf{F-score } \\\hline
only day of week (S)    & 0.88     & 0.80                    & 1.00                 & 0.89                  \\
only reply status (S)            & 0.75     & 0.75                    & 0.75                 & 0.75                  \\
only time hours (S)              & 0.75     & 0.75                    & 0.75                 & 0.75                  \\
only top loved hated (S)         & 0.75     & 1.00                    & 0.50                 & 0.67                  \\
only comment order (S)           & 0.63     & 0.67                    & 0.50                 & 0.57                  \\
only vote updown is reply (S)    & 0.63     & 0.67                    & 0.50                 & 0.57                  \\\hline
only similarity top (S)          & 0.63     & 1.00                    & 0.25                 & 0.40                  \\
only triggered replies range (S) & 0.63     & 1.00                    & 0.25                 & 0.40                  \\
only is reply to has reply (S)   & 0.50     & 0.50                    & 0.25                 & 0.33                  \\
only similarity (S)              & 0.50     & 0.50                    & 0.25                 & 0.33                  \\
only time (S)                    & 0.50     & 0.50                    & 0.25                 & 0.33                  \\
only total comments (S)          & 0.50     & 0.50                    & 0.25                 & 0.33                  \\
only triggered replies total (S) & 0.50     & 0.50                    & 0.25                 & 0.33                  \\
only vote up/down all (S)         & 0.50     & 0.50                    & 0.25                 & 0.33                  \\
only vote up/down total (S)       & 0.50     & 0.50                    & 0.25                 & 0.33                  \\
\hline
All Unscaled                     & 0.50     & 0.00                    & 0.00                 & 0.00                 \\\hline              
\end{tabular}
\caption{Results for classifying 4 paid trolls vs. 4 non-trolls for individual Scaled (S) feature groups. We train on 314 ``mentioned'' trolls vs. 314 non-trolls.
(The top features are better, as they perform well when used alone.)}
\label{table:tr-ment-eval-bivol-some-features}
\end{table*}

Table \ref{table:tr-ment-eval-bivol-some-features} shows the performance of selected feature groups when used in isolation.
% which allows us to get an insight about the performance of individual features.
We can see that features such as time of posting and votes are among the most important ones; yet, in our previous research, we have found them to be virtually irrelevant for finding ``mentioned'' trolls vs. non-trolls \cite{mihaylov-georgiev-nakov:2015:CoNLL}. 

Table \ref{table:tr-ment-eval-bivol-some-features} also shows that the best score is achieved by the day of the week feature, which confirms our assumption that paid trolls tend to write on working days.
Next come the time-related features, which includes hour-related features and number of comments posted during working hours vs. in the evenings.

\section{Discussion}
\label{sec:discuss}

Recall that our objective in this work was to identify paid opinion manipulation trolls in Internet forums.
Unfortunately, we could not train a classifier to do that directly, as we did not have enough known paid trolls.
% to be able to train a proper classifier. 
%In fact, we only had 15 known paid trolls, and for only four of them we had enough posts. 
%In fact, we have a list of trolls that are known to have been paid, but they are only 15, and we could not build a good classifier using only them due to severe class imbalance.
Thus, we resorted to a simple trick: we considered as trolls those users who were accused of being such by other users. 
The assumption was that some of these ``mentioned'' trolls could have actually been paid. 
%Generally, this looks wrong. 
However, this is much of a witch hunt and despite our good overall results, the training data is not 100\% reliable.
For example, some trolls, whether paid or not, could have accused some non-trolls of being trolls, by mistake or on purpose.

\begin{figure}[h!]
  \centering
  \includegraphics[width=\linewidth]{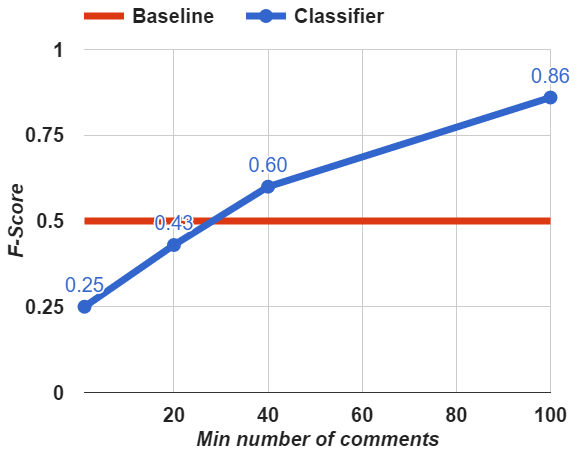}
  \caption{Finding paid trolls with different min number of comments. Training with AS features, and 314 ``mentioned'' trolls vs. 314 non-trolls.}
  \label{figure:paid-trolls-min-comments}
\end{figure}

Recall also that, in our experiments above, we used for testing only four of the fifteen known paid trolls: those with 100 or more comments.
% (for training, we used ``mentioned'' trolls with 150+ comments).
It is interesting to see how our classifier would perform if tested on trolls with different minimum number of comments (and the corresponding number of non-trolls).
This is shown in Figure~\ref{figure:paid-trolls-min-comments}:
we can see that most known paid users with less than 40 comments cannot be exposed as trolls using ``mentioned'' trolls as training examples.

\begin{table}[!]
\begin{tabular}{lrrrr}
\textbf{min mentions}   & \textbf{3}    & \textbf{4}    & \textbf{5}    & \textbf{6}    \\\hline
``mentioned'' trolls          & 536  & 416  & 314  & 259  \\
non-trolls      & 536  & 416  & 314  & 259  \\\hline
%baseline       & 0.50 & 0.50 & 0.50 & 0.50 \\
accuracy       & 0.75 & 0.88 & 0.88 & 0.75 \\\hline
%above baseline & 0.25 & 0.38 & 0.38 & 0.25 \\\hline
F-score        & 0.67 & 0.86 & 0.86 & 0.67\\\hline
\end{tabular}
\caption{
Finding paid trolls with 100+ mentions (4~trolls + 4 non-trolls).
Training with AS features, and users with 150+ comments
and varying minimum number of mentions as a troll.
}
%Experiments with all scaled features, minimum number of (\emph{i})~150 comments and (\emph{ii})~ and different min number of troll mentions for exposed by users trolls and non-trolls and evaluation with 4 known paid trolls with min 100 comments. (balanced)}
\label{table:train-ment-eval-bivol-balanced}
\end{table}

\begin{table}[h]
\begin{tabular}{@{}l@{}rrrr@{}}
\textbf{min mentions}  & \textbf{3}     & \textbf{4}     & \textbf{5}     & \textbf{6}     \\\hline
``mentioned'' trolls              & 536   & 416   & 314   & 259   \\
non-trolls              & 536   & 416   & 314   & 259   \\\hline
accuracy              & 0.83 & 0.87 & 0.91 & 0.92\\\hline
F-score              & 0.83 & 0.87 & 0.91 & 0.92\\\hline
\end{tabular}
\caption{
Finding ``mentioned'' trolls (cross-validation on the training dataset).
Training with AS features, and users with 150+ comments
and varying minimum number of mentions as a troll.
}
\label{table:multiple-min-mentions}
\end{table}

Next, we vary the number of mentions (by different people) needed for us to consider a user a troll; we try 3, 4 and 6, in addition to 5 as above.
Table \ref{table:train-ment-eval-bivol-balanced} shows the results when testing on paid trolls with 100+ mentions (4 trolls + 4 non-trolls), where we trained with All Scaled features, and users with 150+ comments and varying minimum number of mentions as a troll.

Table \ref{table:multiple-min-mentions} shows results when training on the same datasets as in Table \ref{table:train-ment-eval-bivol-balanced}, but this time evaluating with cross-validation on the training data.

We can see that the best results when 
%training with ``mentioned'' trolls and 
testing with paid trolls are achieved for ``mentioned'' trolls with a minimum of 4 or 5 mentions (Table~\ref{table:train-ment-eval-bivol-balanced}), while when both training and evaluating with ``mentioned'' trolls (Table~\ref{table:multiple-min-mentions}), the best results are with 6 mentions. This could mean that paid trolls behave more like moderately ``mentioned'' trolls rather than like highly ``mentioned'' trolls. More experiments, with a higher number of known paid trolls, are needed in order to confirm this.

\begin{figure}[h!]
  \centering
  \includegraphics[width=\linewidth]{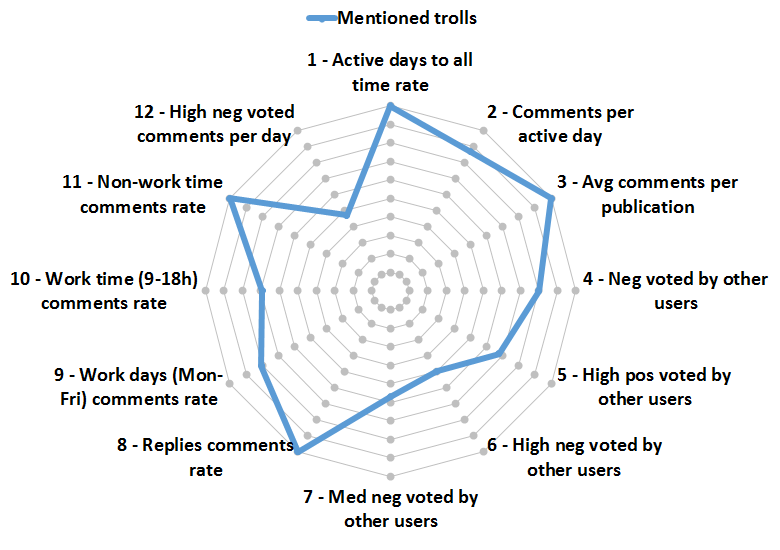}
  \includegraphics[width=\linewidth]{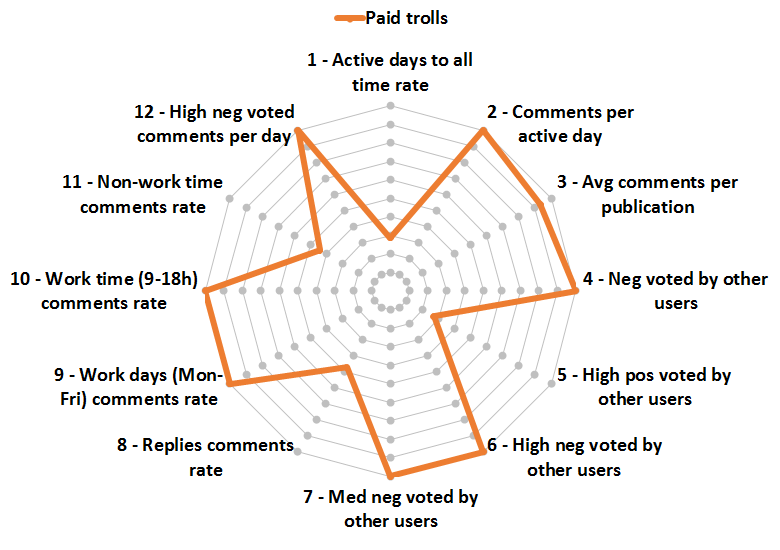}
  \includegraphics[width=\linewidth]{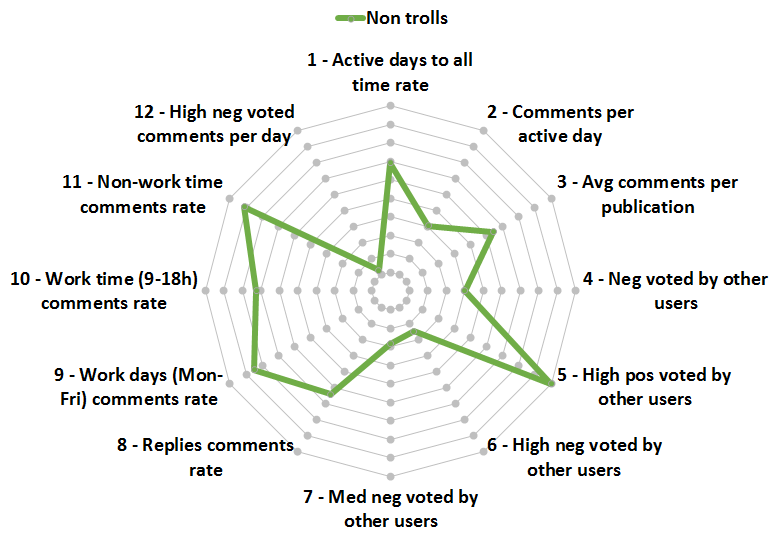}
  \caption{``Mentioned'' trolls vs. paid trolls vs. non-trolls based on average feature values.}
  \label{figure:profile-mentioned-and-paid-and-none}
\end{figure}

%\begin{figure}[h!]
%  \centering
%  \includegraphics[width=\linewidth]{chart-radar-04-profile-mentioned-vs-paid-trolls}
%  \caption{``Mentioned'' trolls and Paid trolls profiles comparison based on average features}
%  \label{figure:profile-mentioned-vs-paid}
%\end{figure}

%\begin{figure}[!tbp]
%  \centering
%  \includegraphics[width=\linewidth]{chart-radar-05-profile-mentioned-vs-paid-vs-non-trolls.png}
%  \caption{``Mentioned'' trolls, Paid trolls and non-trolls profiles comparison based on average features}
%  \label{table:ment-paid-non-feature-comparison}
%  \label{figure:profile-mentioned-vs-paid-vs-none}
%\end{figure}

Finally, we built and analyzed aggregated profiles for the three kinds of users we considered: (\emph{i})~paid trolls vs. (\emph{ii})~``mentioned'' trolls vs. (\emph{iii})~non-trolls.\footnote{Note that we excluded from our analysis users with too few comments or with too few mentions as a troll.}
For this purpose, we selected average values for the most notable features for the users with the highest number of comments from each group.
%These values are shown in Table 11(\ref{table:ment-paid-non-feature-comparison}) for users with the most comments from each user group. It contains some interesting information about the behavior of every user group. 
We then normalized these values with value/max.
The result is shown on Figure~\ref{figure:profile-mentioned-and-paid-and-none}.
%, Figure \ref{figure:profile-mentioned-vs-paid} and Figure \ref{figure:profile-mentioned-vs-paid-vs-none}. 

(1 - Active days to all time rate) shows that ``mentioned'' trolls write at least one comment in 52\% of their days of all time being in the forum, while non-trolls do so 36\% of the time, and paid trolls only do it 15\% of the time. This suggests that paid trolls are less active, maybe because they only write comments when they are paid to do it. 

(2 - Comments per active day) shows that paid trolls and ``mentioned'' trolls write twice as many  comments as non-trolls per day. 

(3 - Avg comments per publication) shows that both paid and ``mentioned'' trolls post more comments per publication than non-trolls. 
%This can be shown on Figure \ref{figure:profile-mentioned-vs-paid-vs-none}. 

(4 - Neg voted by other users), (6 - High neg voted by other users), (7 - Med neg voted by other users) show that both paid and ``mentioned'' trolls have much more negatively voted comments than non-trolls. Yet, this is higher for paid trolls, which could mean that they have more influence compared to the self-driven ``mentioned'' trolls. 

(5) - ``mentioned'' trolls have more positively voted comments compared to paid trolls.% (Figure \ref{figure:profile-mentioned-vs-paid}). 

(8 - Replies comments rate)  - ``mentioned'' trolls are more likely to write comments that are replies to other user's comments compared to non-trolls, while paid trolls prefer to write specific comments and not to enter personal ``battles''. 
%An interesting, but also logical, observation is that 
Moreover, paid trolls are more likely to write comments on working days (9 - Work days (Mon-Fri) comments rate) (Mon-Fri), and during working hours (9-18h) ((10 - Work time (9-18h) comments rate),(11 - Non-work time comments rate)) while ``mentioned'' trolls and non-trolls would write comments at anytime, though mostly during non-working hours.

These observations confirm our assumptions that paid trolls write comments primarily for the money, while ``mentioned'' trolls do so anytime, and are 
%probably 
``self-driven''.
%, inspired by their own beliefs and convictions. 
Yet, note that some of our ``mentioned'' trolls might be actually paid.
%note that we do not know whether some of our ``mentioned'' trolls are not actually paid.

\section{Conclusion and Future Work}

We have presented experiments in trying to distinguish \emph{paid} opinion manipulation trolls vs. non-trolls in Internet forums. As we did not have enough known paid trolls, for training we used ``mentioned'' trolls, assuming that a user who is called a \emph{troll} by several different people is likely to be one, while one who has never been called a troll is unlikely to be such. 
We compared the profiles of (\emph{i})~paid trolls vs. (\emph{ii})~``mentioned'' trolls vs. (\emph{iii})~non-trolls, and we have shown that a classifier trained to distinguish (\emph{ii}) from (\emph{iii}) does quite well also at telling apart (\emph{i}) from (\emph{iii}).
%The evaluation results when testing on actual known paid trolls have shown that we can expose paid trolls with very high accuracy. 
%Yet, this turns out to be true only for trolls with sufficiently high number of posts.

Our further analysis has shown that the most important features were the number of comments, of positive and of negative votes, of posted replies, and the time of commenting.
Overall, paid trolls looked roughly like the ``mentioned'' trolls, except that they were posting most of their comments on working days and during working hours.

Unfortunately, 
%the nature of the features we used here is such that our classifier 
our features
only worked well for trolls with high number of posts. Thus, in future work, we plan to add 
%content features such as 
keywords, topics, named entities, sentiment analysis \cite{sentiBG:RANLP15,sentiMK:RANLP15}, etc, in order to be able to detect ``fresh'' trolls;
this would require
%linguistic processing of the text, e.g., 
stemming \cite{BulStem,Nakov:2003}, POS tagging \cite{Georgiev:2012}, and named entity recognition \cite{georgiev-EtAl:2009:RANLP09}.
We also plan to analyze the comment threads as a whole \cite{barroncedeno-EtAl:2015:ACL-IJCNLP,Joty:EMNLP:2015}.

%The nature of our features means that our paid troll detection works for ``elder trolls'' with at least 150 comments in the forum. In future work, we plan to add content features such as keywords, topics, named entities, etc, in order to detect ``fresh'' trolls.

%\section*{Acknowledgments}
%
%We would like to thank the anonymous reviewers for their constructive comments, 
%which have helped us to improve the paper.

\bibliographystyle{acl}
\bibliography{bib}

\end{document}